\pdfoutput=1

\documentclass[11pt]{article}

\usepackage[preprint]{coling}

\usepackage{xurl}
\usepackage{times}
\usepackage{latexsym}
\usepackage{placeins}
\usepackage{longtable}
\usepackage{multirow}
\usepackage{tabularx}
\usepackage{graphicx} 
\usepackage{lipsum}
\usepackage{adjustbox}

\usepackage[utf8]{inputenc}

\usepackage{microtype}

\usepackage{inconsolata}
\usepackage{float}

\usepackage{booktabs}  
\usepackage{tabularx}  
\usepackage{authblk}

\title{Pragmatic Metacognitive Prompting Improves LLM Performance on Sarcasm Detection}

\author{
  Joshua Lee,
  Wyatt Fong,
  Alexander Le,
  Sur Shah,
  Kevin Han,
  Kevin Zhu\thanks{Corresponding author: \texttt{kevin@algoverse.us}} \\
  Algoverse AI Research
}

\begin{document}
  
\maketitle
  
\begin{abstract}

\begin{figure*}
    \centering
    \includegraphics[scale=0.32]{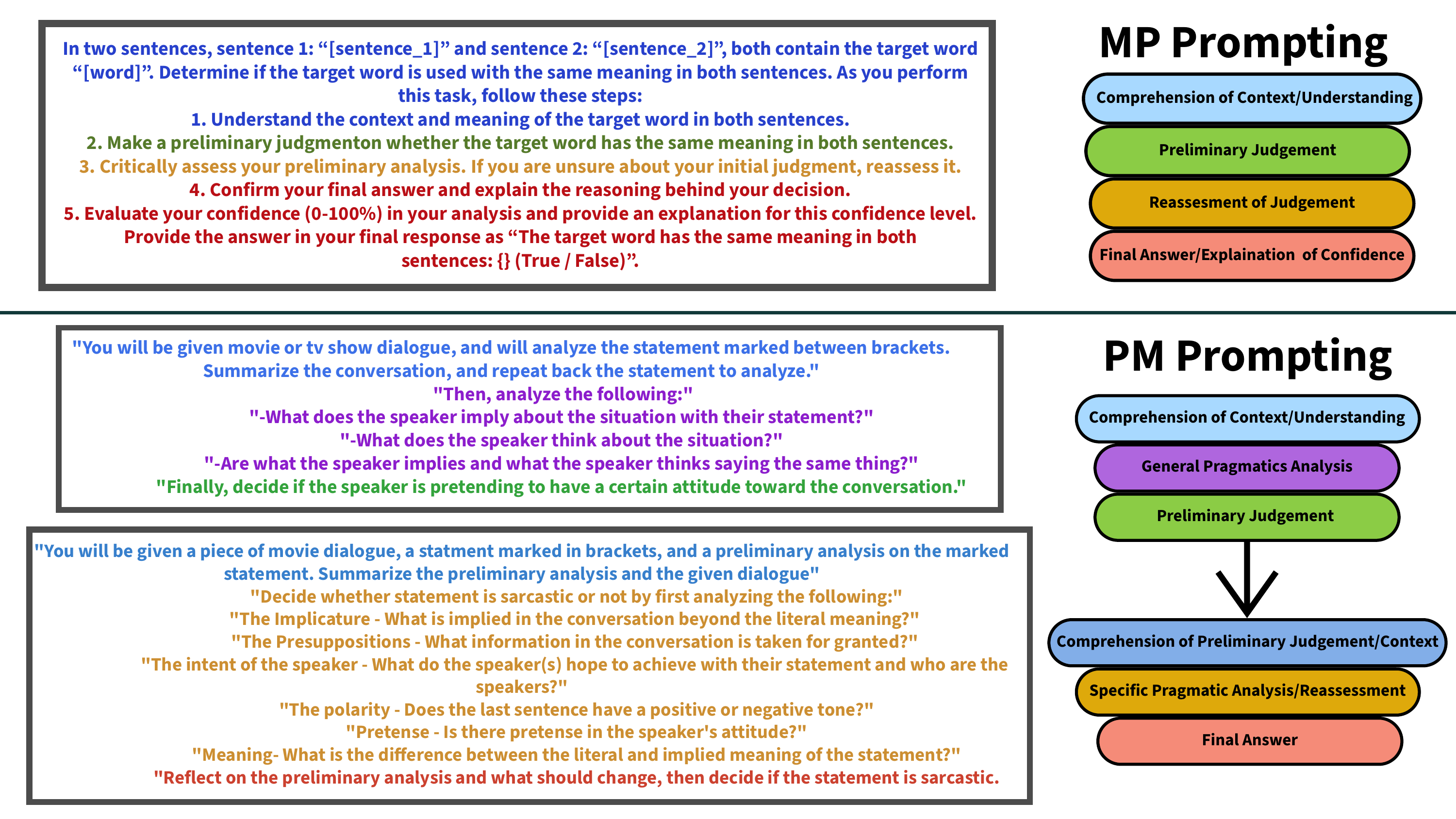}
    \caption{Metacognitive Prompt structure compared to proposed Pragmatic Metacognitive Prompt.}
    \label{fig:pm_prompting-2.pdf}
\end{figure*}

Sarcasm detection is a significant challenge in sentiment analysis due to the nuanced and context-dependent nature of verbiage. We introduce \textbf{Pragmatic Metacognitive Prompting (PMP)} to improve the performance of Large Language Models (LLMs) in sarcasm detection, which leverages principles from pragmatics and reflection helping LLMs interpret implied meanings, consider contextual cues, and reflect on discrepancies to identify sarcasm. Using state-of-the-art LLMs such as LLaMA-3-8B, GPT-4o, and Claude 3.5 Sonnet, PMP achieves state-of-the-art performance on GPT-4o on MUStARD and SemEval2018. This study demonstrates that integrating pragmatic reasoning and metacognitive strategies into prompting significantly enhances LLMs' ability to detect sarcasm, offering a promising direction for future research in sentiment analysis.

\end{abstract}

\section{Introduction}

Within the field of sentiment analysis, various approaches exist to improve emotion classification, from bidirectional transformers to prompt tuning for aspect-based sentiment analysis \citep{Ataei2020, Ouyang2015, devlin2019bertpretrainingdeepbidirectional, li2021sentipromptsentimentknowledgeenhanced, zadeh2017tensorfusionnetworkmultimodal, kanakaraj2015performance}. Yet one present limitation sentiment analysis models face is in determining sarcasm \cite{TanSurvey2023}.

Recent discoveries found that LLMs underperform compared to specially trained transformer encoder models in both sarcasm detection and sentiment analysis. The speculated cause of poor LLM performance is that LLMs are built on logical pipelines, which may contradict sarcasm's non-sequential nature. Regardless, studies believe improving prompting methods is a step towards the solution \cite{Zhang2024, Zhang2023, Tan2023b, Liu2023, Yao2024, Wei2022, Besta2024, Yao2023b}.

This work presents PMP\footnote[1]{Our code can be found at: \url{https://anonymous.4open.science/r/Pragmatic-Metacognitive-Prompting-Improves-LLM-Performance-on-Sarcasm-Detection-C6C2/README.md}} based on \citeauthor{wei2023metacognitive}'s Metacognitive prompting (MP). PMP is a new approach to improving LLM sarcasm detection. Our approach incorporates linguistic principles to mimic how humans reason through emotionally complex text as well as reflection strategies commonly found in LLM reasoning and planning agents \cite{shinn2023reflexionlanguageagentsverbal}. 
This paper presents a novel prompting approach through the use of pragmatics and reflection to improve sarcasm detection, runs its prompting method on sarcasm benchmarks, and at times exceeds the prompt results of the current state-of-the-art (SoTA) prompt for LLM sarcasm detection. 

\section{Background} 

\subsection{Pragmatics}
Pragmatics is a field of linguistics that goes beyond the literal meaning of a conversation. It's the social context of a statement that is needed to comprehend the subtleties of human language.
\cite{Grice1975, Clark1996, Horn2004}. Various studies in linguistics have been conducted on the pragmatics of sarcasm. One pragmatic theory called Grice's Maxims of Conversation, poses the 4 different factors that a conversation must have to be a meaningful conversation. One study, in the field of pragmatics, analyzed Grice's Maxims. It concluded that if Grice's Maxims were exceeded, like with sarcasm in TV shows, it could be a determining factor as to whether dialogue is sarcastic \cite{al2021pragmatic}. 

Our method, PMP, incorporates proposed pragmatic theories on how to detect sarcasm from the field of linguistics into LLM prompting. Our method encourages the LLM to analyze multiple pragmatic theories analyzing sarcasm before reaching its conclusion. A simplified explanation of the theories used in PMP is provided below.

\textbf{The Standard Pragmatic Model (TSPM):} Building upon the foundations of the TSPM, \citeauthor{gibbs2007irony}, refined the understanding of sarcasm detection. His version of TSPM emphasized a process of contacting literal and non-literal meanings alongside context to determine sarcasm.

\textbf{The Pretense Theory of Irony:} When a speaker is ironic or sarcastic, set up a facade to what they actually believe.  \cite{clark1984pretense}. For instance, if someone says ``Your jacket looks soooo nice" in a sarcastic tone, they are presenting an attitude that they do like the look of your jacket when they actually do not.

\textbf{The Echoic Reminder Theory of Verbal Irony:} 
This method is characterized by positive and neutral statements that ironically reference a past statement. It is often used as a critique of a scenario. An example would be the phrase ``What a great idea!" which typically has a positive connotation. However, if it was used to describe a terrible plan it would take a negative connotation, conveying sarcasm. \cite{kreuz1989how}.

\subsubsection{Implicature} 
Implicatures are implicit inferences drawn automatically from the information provided in a sentence, relying on the shared context between the speaker and listener.  For example, if Bob says “Do you want any cake for lunch?” and Joey responds “I don’t want to get fat”, the implied meaning is that Joey is declining the offer. However, inferences like these can sometimes be incorrect. For instance, Joey may have been making a completely unrelated comment. As the interpreters, we may assume that Joey's statement was relevant to the conversation, which causes us to infer his refusal. 

\subsubsection{Presuppositon} 
The presupposition is information automatically accepted as true in order for a statement to make sense. For example, in the statement ``The king of France is bald", the presupposition is that there exists a King of France. This statement assumes that a king exists in France, even though there might not be one for the sake of making sense of the statement.

\begin{table*}[!ht]
\begin{center}
\begin{tabular}{c c c c c}
\toprule 
    \multicolumn{1}{c}{} & \multicolumn{2}{c}{SemEval 2018} & \multicolumn{2}{c}{MUStARD} \\
    \midrule
      Model & Acc. & Ma-F1 & Acc. & Ma-F1 \\
      \midrule
        \multirow{1}{*}{GPT-4o (IO)} 
       & 64.03 & 63.17 & 67.24& 65.79\\
        \multirow{1}{*}{GPT-4o (CoT)} 
        & 58.92 & 51.99 & 58.11 & 55.76\\
        \multirow{1}{*}{GPT-4o (ToT)} 
       & 63.90 & 63.02 & 69.00 & 68.27\\
        \multirow{1}{*}{GPT-4o (CoC)} 
        & 70.79 & 70.60 & 69.42 & 68.48\\
        \multirow{1}{*}{GPT-4o (GoC)} 
      & 74.03 & 74.02 & 70.69 & 69.91 \\
        \multirow{1}{*}{GPT-4o (BoC)} 
        & 62.12 & 61.85 & 69.42 & 68.45\\
        \multirow{1}{*}{GPT-4o (PMP)} 
       & \textbf{86.68} & \textbf{83.18}  & \textbf{79.42}& \textbf{77.65}  \\
        \multirow{1}{*}{GPT-4o-mini (PMP)} 
       & 81.88 & 79.85 & 65.79 & 62.29\\
        \midrule
        \multirow{1}{*}{Claude 3.5 Sonnet (IO)} 
       & 75.13 & 75.11 & \textbf{74.78} & \textbf{74.78}\\
        \multirow{1}{*}{Claude 3.5 Sonnet (CoT)} 
       & 71.56 & 71.47 & 73.62 & 73.53 \\
        \multirow{1}{*}{Claude 3.5 Sonnet (ToT)} 
       & 68.62 & 68.61 & 58.84 & 54.46\\
        \multirow{1}{*}{Claude 3.5 Sonnet (CoC)} 
       & \textbf{82.27} & \textbf{82.23} & 74.20 & 74.16\\
        \multirow{1}{*}{Claude 3.5 Sonnet (GoC)} 
        & 57.33 & 57.24 & 52.77 & 52.67\\
        \multirow{1}{*}{Claude 3.5 Sonnet (BoC)} 
        & 65.94 & 65.50 & 59.71 & 56.70\\
         \multirow{1}{*}{Claude 3.5 Sonnet (PMP)} 
       & 81.50 & 76.72 & 72.60 & 71.66\\
       \midrule
        \multirow{1}{*}{LLaMA-3-70B (PMP)} 
        & 80.86 & 78.15 & 72.73 & 73.06\\
        \midrule
        \multirow{1}{*}{LLaMA-3-8B (IO)} 
       & 49.36 & 44.47 & 54.64 & 44.99\\
        \multirow{1}{*}{LLaMA-3-8B (CoT)} 
       & 49.36 & 44.55 & 54.20 & 44.86\\
        \multirow{1}{*}{LLaMA-3-8B (ToT)} 
        & 50.64 & 48.63 & 54.35 & 50.56\\
        \multirow{1}{*}{LLaMA-3-8B (CoC)} 
        & 49.23 & 44.36 & 54.93 & 45.66\\
        \multirow{1}{*}{LLaMA-3-8B (GoC)} 
       & 57.33 & 57.24 & 52.7 & 52.67\\
        \multirow{1}{*}{LLaMA-3-8B (BoC)} 
        & 65.94 & 65.50 & 59.71 & 56.70\\
        \multirow{1}{*}{LLaMA-3-8B (ToC)} 
        & 68.88 & 68.21 & \textbf{61.26} & \textbf{58.03}\\
        \multirow{1}{*}{LLaMA-3-8B (PMP)} 
        & \textbf{78.21} & \textbf{77.65} & 53.48 & 54.69\\
        
    \bottomrule
    \end{tabular}
    \caption{Comparison of PMP with Claude 3.5 Sonnet, GPT4o, GPT4o-mini, LLaMa-3-70B and LLaMA-3-8B to prompting methods. The best results are bolded.}
    \end{center}
    \label{fig:table2}
\end{table*}

\section{Method}

\subsection{Design}
Our prompting method builds on top of \citeauthor{wei2023metacognitive}'s Metacognitive prompting (MP). MP consists of prompting an LLM to repeat the given information, create a preliminary analysis, reflect on their preliminary analysis, and then create a final judgment \cite{wei2023metacognitive}. See Figure~\ref{fig:PM prompting-2.pdf} for more details on MP.
In our method, PMP, the LLM is encouraged to analyze simplified elements of pragmatic theories in the preliminary analysis and reflection stages. We establish two separate LLM calls, one which analyzes the prompt from the lens of each pragmatic factor: implicature, presuppositions, intent, polarity, pretense, and potential meanings individually, and a second LLM call that reflects on the analysis and outputs a final prediction. A detailed explanation of PMP is provided in Figure~\ref{fig:PM prompting-2.pdf}.

\section{Experimental Design}

\subsection{Benchmarks}
We evaluated our sarcasm detection method on the same benchmarks as \cite{Yao2024}: MUStARD \cite{castro-2019}, which consists of sarcastic and non-sarcastic comments in TV and movie dialogue paired with context; and SemEval 2018 Task 3 \cite{seminar-2018} consisting of sarcastic and non-sarcastic twitter statements.

\subsection{Models}
We tested our method using models also utilized in SarcasmCue \cite{Yao2024}. The models are: GPT-4o, LLaMA 3-8B and Claude 3.5 Sonnet \cite{anthropic_claude_models}. Furthermore, we additionally tested on GPT-4o mini \cite{openai2023gpt4} and LLaMA 3-70B \cite{touvron2023llama}.

\subsection{Baselines}
\subsubsection{SarcasmCue}
Our method achieves the new SoTA in comparison to SarcasmCue. The method SarcasmCue modifies popular SoTA prompts to analyze a ``cue'', which is a coherent language sequence that serves as an indicator towards identifying sarcasm, from either linguistic (rhetorical devices, punctuation), contextual (topic, common knowledge), or emotional (emotional words, emojis) parts of a sentence.

SarcasmCue\footnote[2]{At the time of the writing of this paper, \citeauthor{Yao2024} has not published their source code. Therefore, we compare our results with the reported results from their paper.} introduces four sarcasm detection methods; three prompting techniques: Bag of Cues (BoC), Chain of Cues (CoC), and Graph of Cues (GoC), and one that requires explicit model training, Tensor of Cues (ToC). BoC removes sequential bias by treating cues independently. CoC arranges cues in a sequential order to capture the step-by-step reasoning process of sarcasm detection. GoC analyzes the relationships between cues without imposing a fixed sequence. ToC adds encoded indications through explicit training to leverage higher-order interactions among cues. For the exact prompts, please see Appendix \ref{sec:prompts}.


\section{Results}
The accuracy and Macro-F1 scores comparing PMP with prompting method baselines are compared in Tables 1. The accuracy and Macro-F1 scores comparing PMP with SarcasmCue's BoC, CoC, GoC, and ToC strategies are reported in Table 1. 

\textbf{Comparison to Popular Prompting methods:} PMP surpasses popular prompting methods such as Zero Shot, Chain of Thought, and Tree of Thought in both SemEval 2018 Task 3 and MUStARD. Across both performing well on LLaMA-3-8B and GPT-4o, with the exception of Claude 3.5 Sonnet. Zero-shot prompting still works well with Claude 3.5 Sonnet in 2 benchmarks, aligning with \citeauthor{Yao2024}'s results. PMP's performance with LLaMA-3-70B is significantly higher than with LLaMA-3-8B.

\textbf{Comparison to State of the Art (SoTA):} PMP is competitive with and exceeds SarcasmCue's performance on all datasets with GPT-4o, while performing well on LLaMA-3-8B on SemEval 2018. As shown in Table 2, zero-shot Claude 3.5 Sonnet achieves the highest accuracy on the MUStARD datasets, outperforming its performance in SarcasmCue and PMP. 

\textbf{Datasets:} PMP performs the best on SemEval 2018 Task 3, although it falls slightly short of SarcasmCue on Claude. PMP struggles on Sarcasm Corpus V1 the most, with current SoTA and Tree of Thought outperforming it across certain models.

\textbf{State of The Art and PMP:} Between PMP and SarcasmCue, neither consistently achieve higher accuracies than the latter across all models and datasets, excluding Claude. However, one notable factor is that for both datasets, GPT-4o utilizing PMP performs best in comparison with all other models and prompting methods. 
GPT-4o outperforming other LLMs aligns with previous studies such as \citeauthor{Zhang2024}'s work, suggesting GPT4o's performance is a common factor in sarcasm detection. Another inconsistency is that SarcasmCue underperforms some prompts in SemEval 2018 Task 3 across all models except Claude 3.5 Sonnet, while PMP outperforms prompts in SemEval 2018 Task 3 across all models but underperforms SoTA in Claude 3.5 Sonnet. Analyzing SemEval 2018 as a dataset could help explain these performance patterns. 

\section{Conclusion}
Pragmatic Metacognitive Prompting is a novel approach for enhancing sarcasm detection in LLMs. PMP is competitive with or beats the current state-of-the-art methods for sarcasm detection with pretrained LLMs such as GPT4o and LLaMA-3-8B. It introduces various pragmatic theories into the prompt design, fosters a deeper contextual understanding that improves sarcasm identification, and incorporates a human-like reflection step for final verification and sarcasm reasoning. After testing across models like GPT-4o and LLaMA-3-8B, PMP underscores the potential of pragmatic-informed methods to outperform traditional prompting methods, and point to a continued focus on linguistic theories to bridge performance gaps in sentiment analysis.

\section{Limitations}
While PMP represents on approach to implementing pragmatic reflection, prompting is only one implementation of pragmatics and reflection in natural language processing. A key limitation to using zero shot prompting is that PMP does not guarantee high performance in sarcasm detection that deviates from general linguistics norms and in domain-specific contexts. Due to PMP's reliance on LLM's pretraining with data, underrepresented cultural or lingustic norms are also not accounted for with prompting. These limitations suggests PMP is a step towards improving sarcasm detection, but does not represent a comprehensive solution. 

\bibliography{paper}

\appendix
\section{Prompts}
\label{sec:prompts}

The prompting method we utilized in our approach guides the model through a structured reasoning process before reaching a conclusion. The prompt instructs the model to analyze a statement checking its comprehension on the given information, before asking the LLM to generate an accompanying preliminary analysis analyzing basic pragmatic factors. After completing the preliminary analysis, the model then passes its generated analysis to another LLM call, where the model has a chance to reflect and comprehend the preliminary analysis originally generated. It then directs the model to refine the analysis by systematically addressing specific pragmatic aspects, including implicature, presuppositions, speaker intent, polarity, pretense, and the relationship between literal and implied meanings. The wording of our initial prompt varies per dataset to ensure that all information about the benchmark is given for the LLM to generate a proper analysis. An example of this would be with MUStARD's dataset:\\ 

``You will be given movie or TV show dialogue and will analyze the statement marked between brackets. Summarize the conversation, and repeat back the statement to analyze."
``Then, analyze the following:"

Decide whether the statement is sarcastic or not by first analyzing the following:
\begin{enumerate}
    \item The Implicature – What is implied in the conversation beyond the literal meaning?
    \item The Presuppositions – What information in the conversation is taken for granted?
    \item The Intent of the Speaker – What do the speaker(s) hope to achieve with their statement, and who are the speakers?
    \item The Polarity – Does the last sentence have a positive or negative tone?
    \item Pretense – Is there pretense in the speaker's attitude?
    \item Meaning – What is the difference between the literal and implied meaning of the statement?//
\end{enumerate}
Reflect on the preliminary analysis and what should change, then decide if the statement is sarcastic." \\
\subsection{Cues}
\subsubsection{Bag of Cues}
The Bag of Cues method evaluates sarcasm by treating cues and without order.

\textbf{Prompt Example:}
``Identify if the given statement is sarcastic based on the presence of the following cues:

Rhetorical devices (e.g., irony, hyperbole, or understatement)
Emotional language (e.g., frustration, happiness, or sarcasm-laden phrases)
Contextual inconsistencies (e.g., contradictory meanings or unexpected word choices).
Does the statement exhibit any of these cues?"

\textbf{Example Application:}
Input: ``Oh, great! Another meeting that could have been an email."
Rhetorical Device: Irony
Emotional Cue: Frustration
Contextual Cue: Work-related sarcasm
Detection: Likely sarcastic

\subsubsection{Chain of Cues}
The Chain of Cues method evaluates sarcasm by analyzing cues sequentially. It simulates logical reasoning to then assess the overall sarcastic nature of a statement.

\textbf{Prompt Example:}
``Analyze the statement step-by-step: Identify any rhetorical device (e.g., hyperbole, irony). Determine if emotional cues such as frustration or humor are present. Check for contextual markers that may suggest sarcasm. Does the progression or order  of these cues indicate sarcasm?"

\textbf{Example Application:}
Input: ``Thanks for breaking the printer. Really helpful."
Rhetorical Device: Irony detected in ``Really helpful."
Emotional Cue: Frustration in the context of the statement.
Contextual Marker: Complaints about a broken printer.
Detection: Sarcastic

\subsubsection{Graph of Cues}
The Graph of Cues method evaluates sarcasm by analyzing the relationships between cues. This method leverages interdependencies between linguistic, emotional, and contextual features.

\textbf{Prompt Example:}
``Construct a graph where:
Nodes represent sarcasm cues (e.g., rhetorical devices, emotional cues, contextual features).
Edges represent relationships between these cues (e.g., reinforcement, contrast).
Based on the interconnected cues, does the statement appear sarcastic?"

\textbf{Example Application}:
Input: ``Wow, you’re so good at driving (said during a near accident)."
Nodes:
Rhetorical Device: Sarcastic praise (``so good").
Emotional Cue: Anxiety/frustration.
Contextual Cue: Near accident.
Edges:
Reinforcement between rhetorical device and emotional cue.
Contextual cue amplifies sarcasm.
Detection: Sarcastic
\subsubsection{Tensor of Cues}
The Tensor of Cues method uses a structured, multi-dimensional representation of sarcasm cues to train a model explicitly. This approach captures interactions between cues in a numerical format.
The implementation details include implementation details
linguistic, emotional, and contextual cues to be encoded as tensors. The model is then trained to revolve around optimizing the model to learn patterns across these dimensions.

\textbf{Example Tensor Encoding:}
Input: ``Nice job ignoring me all day!"
Linguistic Cue: Irony (tensor dimension 1).
Emotional Cue: Frustration (tensor dimension 2).
Contextual Cue: Social neglect (tensor dimension 3).
Combined Tensor: Captures interrelations of cues for sarcasm prediction.

\textbf{Performance Highlights:}
This method achieves higher accuracy by explicitly modeling multi-cue interactions compared to the prompting methods.
This structured prompt ensures that the model’s reasoning aligns with pragmatic analysis principles, fostering a more nuanced understanding of sarcasm detection.

\section{LLM Pragmatic Reasoning}
\begin{figure*}
    \centering
    \includegraphics[scale=0.32]{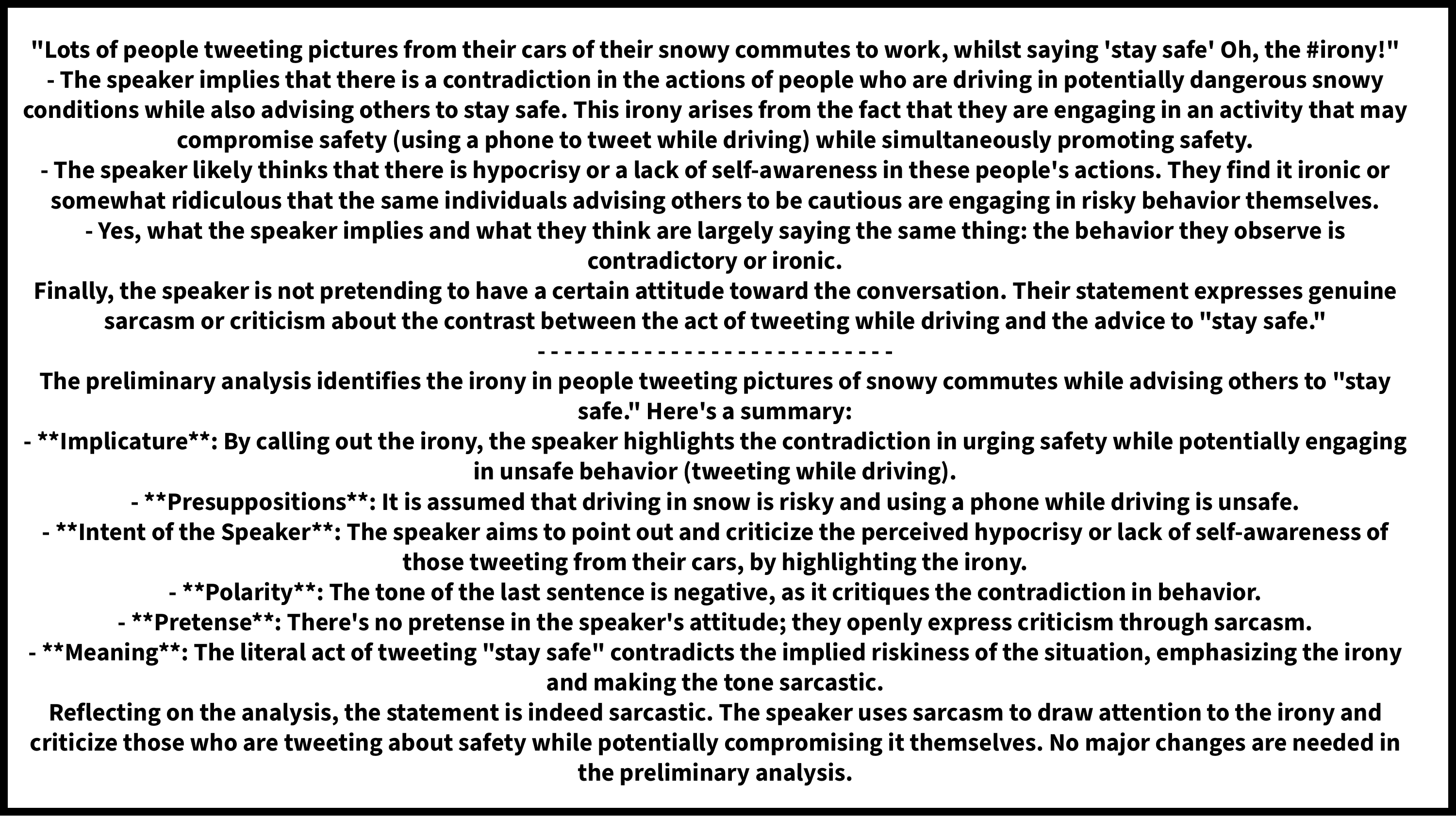}
    \caption{PMP analysis on the phrase ``Lots of people tweeting pictures from their cars of their snowy commutes to work, whilst saying 'stay safe' Oh, the \#irony!".}
    \label{fig:prompt_example_6.pdf}
\end{figure*}

Figure \ref{fig:prompt_example_6.pdf} illustrates the application of PMP and the reasoning process demonstrated by the model. As depicted, the LLM leverages various elements of the pragmatic framework to arrive at a well-considered conclusion. In the appendix, we include a detailed PMP analysis of the phrase, \textit{“Lots of people tweeting pictures from their cars of their snowy commutes to work, whilst saying ’stay safe’ Oh, the \#irony!”}. In this example, PMP successfully identifies the nuanced contrast between literal and implied meanings, allowing the model to detect sarcasm effectively by contextualizing the speaker's intent, polarity, and presuppositions.

\end{document}